\documentclass[sigconf]{acmart}
\usepackage{stfloats}
\AtBeginDocument{%
  \providecommand\BibTeX{{%
    \normalfont B\kern-0.5em{\scshape i\kern-0.25em b}\kern-0.8em\TeX}}}

\setcopyright{none}
\renewcommand\footnotetextcopyrightpermission[1]{}

\acmConference[Woodstock '18]{Woodstock '18: ACM Symposium on Neural
  Gaze Detection}{June 03--05, 2018}{Woodstock, NY}
\acmBooktitle{Woodstock '18: ACM Symposium on Neural Gaze Detection,
  June 03--05, 2018, Woodstock, NY}
\acmPrice{15.00}
\acmISBN{978-1-4503-XXXX-X/18/06}

\begin{document}
\title{Learning to segment from misaligned and partial labels}
\pagestyle{empty}
% \acmcopyrightmode{0}
\author{Simone Fobi}

\authornotemark[1]
\affiliation{%
  \institution{Columbia University}
%   \city{New York City}
%   \state{NY}
}
\email{sf2786@columbia.edu}

\author{Terence Conlon}
\affiliation{%
  \institution{Columbia University}}
\email{tmc2180@columbia.edu}

\author{Jayant Taneja}
\affiliation{%
  \institution{University of Massachusetts Amherst}
}
\email{jtaneja@umass.edu}

\author{Vijay Modi}
\affiliation{%
 \institution{Columbia University}}
 \email{modi@columbia.edu}
 %\streetaddress{Rono-Hills}
 %\city{Doimukh}
 %\state{Arunachal Pradesh}
 %\country{India}}

% \affiliation{%

% \affiliation{%



\renewcommand{\shortauthors}{Fobi, et al.}

\begin{abstract}
To extract information at scale, researchers increasingly apply semantic segmentation techniques to remotely-sensed imagery. While fully-supervised learning enables accurate pixel-wise segmentation, compiling the exhaustive datasets required is often prohibitively expensive. As a result, many non-urban settings lack the ground-truth needed for accurate segmentation. Existing open source infrastructure data for these regions can be inexact and non-exhaustive. Open source infrastructure annotations like OpenStreetMaps are representative of this issue: while OpenStreetMaps labels provide global insights to road and building footprints, noisy and partial annotations limit the performance of segmentation algorithms that learn from them.

In this paper, we present a novel and generalizable two-stage framework that enables improved pixel-wise image segmentation given misaligned and missing annotations. First, we introduce the Alignment Correction Network to rectify incorrectly registered open source labels. Next, we demonstrate a segmentation model -- the Pointer Segmentation Network -- that uses corrected labels to predict infrastructure footprints despite missing annotations. We test sequential performance on the Aerial Imagery for Roof Segmentation dataset, achieving a mean intersection-over-union score of 0.79; more importantly, model performance remains stable as we decrease the fraction of annotations present. We demonstrate the transferability of our method to lower quality data sources, by applying the Alignment Correction Network to OpenStreetMaps labels to correct building footprints; we also demonstrate the accuracy of the Pointer Segmentation Network in predicting cropland boundaries in California from medium resolution data. Overall, our methodology is robust for multiple applications with varied amounts of training data present, thus offering a method to extract reliable information from noisy, partial data. 
\end{abstract}






\maketitle

\section{Introduction}
\label{introduction}
Processing remotely-sensed imagery is a promising approach to evaluate ground conditions at scale for little cost. Algorithms that intake satellite imagery have accurately measured crop type \cite{rustowicz2019},\cite{kussul2019}, cropped area \cite{du2019}, building coverage \cite{xu2018} \cite{geoseg2019}, urbanization \cite{alshehhi2017}, and road networks \cite{cadamuro2019} \cite{xu2018roads}. However, successful implementation of image segmentation algorithms for remote sensing applications depends on large amounts of data and high-quality annotations. Wealthy, urbanized settings can more readily apply segmentation algorithms, due to either the presence of or the ability to collect significant amounts of carefully annotated data. In contrast, more rural regions often lack the means to exhaustively collect ground truth data. Some open source datasets exist for such settings, and by successfully coupling these annotations with remotely sensed imagery, researchers can gain insights into the status of infrastructure and development where well-curated sources of these data do not exist. \cite{kaiser2017} \cite{audebert2017}. \newline
\begin{figure}[t]
\vskip 0.2in
\begin{center}
\centerline{\includegraphics[width=\columnwidth]{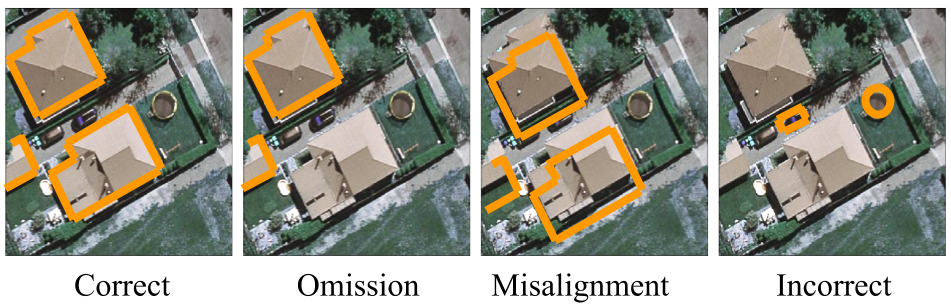}}
\caption{Types of label noise present in open source data. Building footprints are the class of interest.}
\label{noisylabels}
\end{center}
\vskip -0.2in
\end{figure}
Although these global open source ground truth datasets -- e.g. OpenStreetMaps (OSM) -- offer large amounts of labels for use at no cost, the annotations within suffer from multiple types of noise \cite{mnih2012} \cite{basiri2016}: \textit{missing or omitted annotations}, defined as objects being present in the image and not existing in the label \cite{mnih2012}; \textit{misaligned annotations} occur when annotations are translated and/or rotated from its true position \cite{vargas2019}; and \textit{incorrect annotations} -- annotations that do not directly correspond to the object of interest in the image. Figure \ref{noisylabels} presents examples of these three types of label noise. 

Noisy datasets present a training challenge when using traditional segmentation algorithms, as the model cannot learn to associate image features and target labels when the relationship is obscured by noise. To address the issues of misaligned and omitted annotations, and in order to extract information from imperfect data, we present a simple and generalizable method for pixel-wise image segmentation. First, we address annotation misalignment by proposing an Alignment Correction Network (ACN). With a small number of images and human verified ground truth annotations, the ACN learns to correct misaligned labels. Next, the corrected open source annotations are used to train the Pointer Segmentation Network (PSN), a model which takes in a point location and identifies the object containing that point. Learning associations from a representative point is a widely acknowledged method of object detection: \cite{fei2016} notes that an intuitive way for humans to refer to an object is through the action of pointing. By `\textit{pointing-out}' the object instance of interest, our network ignores other instances that may not have corresponding annotations, therefore preventing performance degradation caused by annotation-less instances within the image. As a result, our sequential approach presents a method for handling misaligned data as well as varying levels of label completeness without explicitly changing the loss function to compensate for noise. While our approach cannot replace large amounts of carefully annotated outlines, it can complement existing open source datasets and algorithms, reduce the cost of obtaining large amounts of full annotations, and allow researchers to extract information from imperfect datasets.
This paper's key contributions are as follows:
\begin{itemize}
    \item We introduce the Alignment Correction Network (ACN), a means to verify and correct misaligned annotations using a small amount of human verified ground truth labeled data.
    \item We propose the Pointer Segmentation Network (PSN), a model that can reliably predict polygon boundaries on remotely-sensed imagery despite omitted training annotations and without requiring any bespoke loss functions.
    \item We demonstrate the applicability of our methodology to three different segmentation problems: building footprint detection with a highly-accurate dataset, building footprint detection with noisier training data, and cropland boundary prediction.
\end{itemize}
Taken as a whole, our approach enables resource constrained actors to use large amounts of misaligned and partial labels -- coupled with a very small amount of human verified ground truth annotations -- to train image segmentation algorithms for a variety of tasks. 
The rest of the paper is organized as follows: In \textit{Related Work}, we discuss related literature; in \textit{Methods}, we describe our novel methodological contributions; in \textit{Results}, we present results for the ACN and the PSN for all segmentation tasks; and in \textit{Conclusion}, we restate our most salient findings. 

\begin{figure*}
  \centering
  \includegraphics[width=1.75\columnwidth]{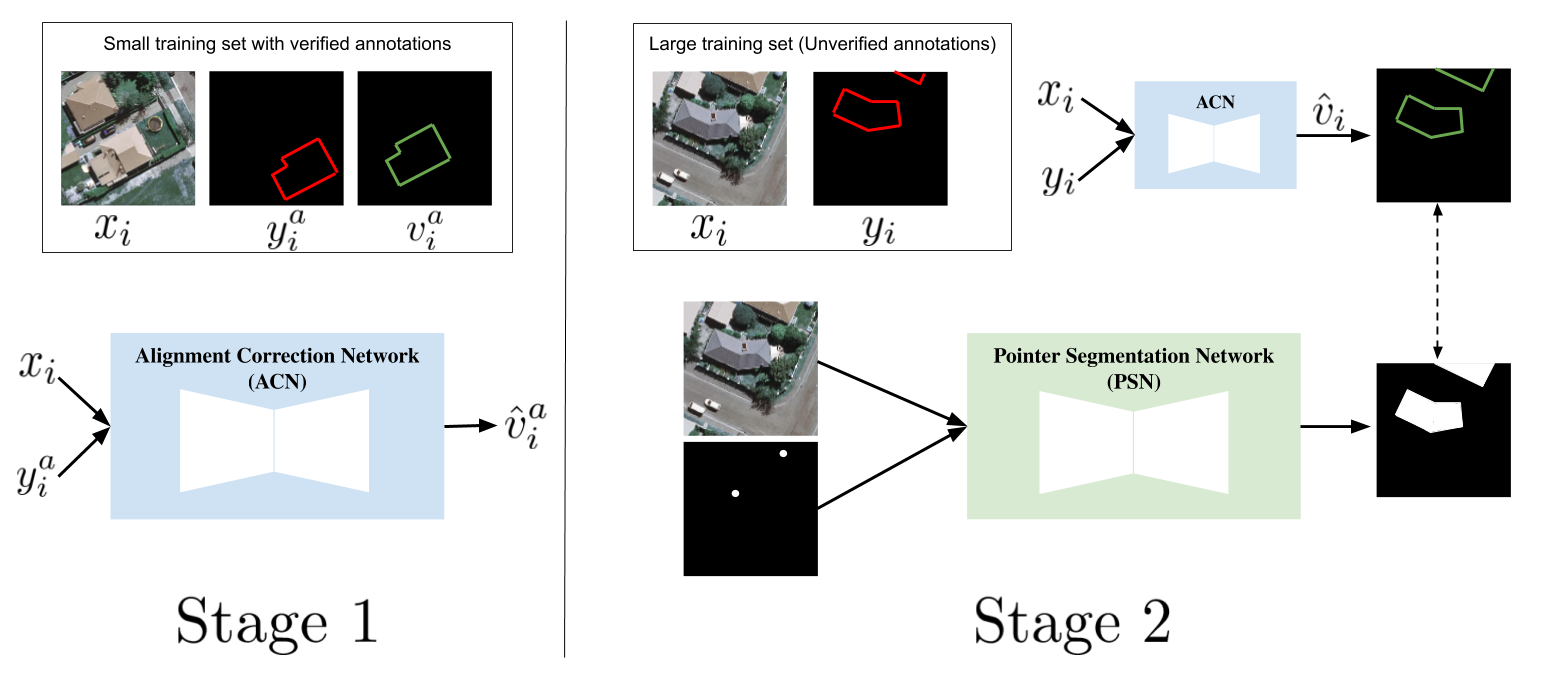}
  \caption{Summary of our two-stage approach to segment from noisy annotations. Stage 1: The ACN uses an image ($x_i$) and label ($y_i^a$) with a single misaligned annotation to predict a corrected annotation $\hat{v}_{i}^{a}$ containing the realigned annotation. Random shifts between $\pm$10 pixels are applied to $v_i^a$ to obtain $y_i^a$. The network is trained with a small set of images ($x$) and verified ground truth annotations ($v$).  Stage 2: A large noisy training set is first realigned with the ACN. Realigned, incomplete annotations are used for supervision. The PSN uses selected points from available instances, $x_i$ and $\hat{v}_{i}$ to learn the segmentation task.}~\label{fig:architecture}
\end{figure*}

\section{Related Work}
\label{related}
Computer vision researchers have recently made numerous advances in semantic segmentation, in applying state-of-the art techniques to remote sensed imagery, and in learning from noisy datasets; we discuss some important contributions to the literature below.

\textbf{Existing Segmentation Approaches} \\
Primarily based on improvements to deep convolutional neural networks (DCNN) architectures, researchers have achieved record performances for a variety of different segmentation tasks. Fully convolutional encoder-decoder type architectures -- one type of DCNN -- take in an image and output a per-pixel prediction for the class of interest \cite{long2015}. Some architectures use symmetric networks with skip connections to perform pixel-wise predictions \cite{ronnerberger2015} \cite{segnet2017}. Alternatively, two-stage detection algorithms first perform region proposal -- areas that have a high likelihood of containing the object of interest -- and then detect objects within the identified regions  \cite{girshick2014} \cite{girshick2015} \cite{ren2015}. Modifications to two-stage detection algorithms have enabled semantic segmentation of images, whereby individual pixels in an image are placed into one of a number of classes \cite{he2017} \cite{li2017}. Development of these segmentation architectures has been facilitated by large, comprehensive datasets which enable the implementation of these algorithms in a fully supervised approach: here, every object in the image and its corresponding annotation are used in the learning process \cite{pascal-voc-2012} \cite{MartinFTM01} \cite{mscoco}.

\textbf{Applying Deep Learning to Remote Sensed Imagery} \\
Multiple projects have leveraged satellite imagery to answer various questions on land use, road quality, object detection, consumption expenditure: by linking sparse ground truth with abundant imagery, researchers can extrapolate trends in existing data to areas where labeled data do not exist \cite{safyan}, \cite{mappingfromnl}, \cite{bikash_autorooftop}. Alternatively, some works have proposed neural network architectures that sidestep training data constraints and the relative lack of labeled ground-truth in remote areas \cite{jiaqi_improvedseg} \cite{povertymapping2}. Jean et al. combine Google maps daytime images (provided by DigitalGlobe), nighttime lighting, and survey data to estimate poverty for multiple African countries \cite{povertymapping}. High resolution daytime images were used to train a model to predict nighttime lights as measured by DMSP-OLS; features extracted from the last layer of the model were then used to estimate household expenditure or wealth. Results from this paper suggest that predictions about economic development can be made from remote sensed data using features derived from imagery; this insight provides additional motivation for developing methods that extract information from noisy imagery datasets.

\textbf{Learning From Noisy Annotations} \\
The problem of poor-quality training data, especially in rural areas, for segmentation tasks is well known:  \cite{mahabir2017authoritative} acknowledge the variability in coverage of open source data in Kenya and observe significant degradation of coverage as one moves away from urban settings. Coverage degradation from urban to rural areas is also seen in South Africa\cite{siebritz2014assessing}, Brazil\cite{camboim2015investigation} and Botswana\cite{botswana_complete}. \cite{swan_noise} estimates the effects of multiple types of training data noise, including misalignment and missing annotations, finding that as noise levels increase, both precision and recall decrease. For applications such as measuring building or field area which are useful in downstream analysis of wealth, crop yield and more, high noise levels decrease the ability to successfully use segmentation algorithms.  
Several works tackle the problem of learning from imperfect labels. \cite{mnih2012} propose new loss functions to address noisy labels in aerial images. \cite{vargas2019} \cite{girard2019noisy} both focus on the issue of misalignment: \cite{girard2019noisy} uses a self-supervised approach to align cadaster maps, and while the method proposed in \cite{vargas2019} maximizes the correlation between annotations and outputs from a building prediction CNN, it assumes buildings in small groups have the same alignment error. Our two-stage approach builds upon existing convolutional frameworks common to many noise correction approaches. However our approach relies on the well-known binary cross entropy loss function, addresses both misalignment and omitted annotation, and does not require that all misalignments are identical. Thus serving as an attractive alternative when noisy labels are present.

\section{Methods}
\label{methods}
Traditional segmentation methods take an image input $x_i$ and aim to learn a function $f(\textbf{x})$ that predicts a single channel label $\hat{\upsilon}_{i}$ containing all building instances present in the image. Equation \ref{eq:baseline_notation} shows the learned function given $x_i$, where $\upsilon_i^a$ is the single channel label of instance \textit{a} in image $x_i$ and there are a total of A instances in that image:
\begin{equation}\label{eq:baseline_notation} 
\begin{split}
    & f(x_{i}) \rightarrow \hat{\upsilon}_{i} \\
    & s.t.\hspace{0.3cm}\hat{\upsilon}_{i} = \hat{\upsilon}_{i}^{1} \cup \hat{\upsilon}_{i}^{2} ... \cup  \hat{\upsilon}_{i}^{A}\\
\end{split}    
\end{equation}
% In all model training, we implement a binary cross-entropy loss that compares the predicted label $\hat{\upsilon}_{i}$ to the true label $\upsilon_i$. 

\subsection{Alignment Correction Network}
Misalignment occurs when there is a registration difference between an object in an image and its annotation. In remote sensing, misaligned annotations may occur for a number of reasons, including human error and imprecise projections of the image \cite{girard2019noisy}. There are two types of annotation alignment errors: 1) translation errors, where the annotation is shifted relative to the object, and 2) rotation errors, where the annotation is rotated relative to the object. \cite{vargas2019} suggest that translation errors are more frequent for OpenStreetMaps in rural areas. Thus in this paper, we only address translation errors present in open source data. We propose an Alignment Correction Network (ACN) that takes in an image $x_{i}$ and a label $y_{i}^{a}$ containing one misaligned instance \textit{a}. The ACN outputs a label $\hat{\upsilon}_{i}^{a}$ containing the predicted, corrected annotation. $\hat{\upsilon}_{i}^{a}$ is compared to $\upsilon_i^a$ to learn optimal weights for the network. During training, the misaligned label $y_{i}^{a}$ is obtained by applying random x-y shifts, between $\pm$10 pixels to $\upsilon_i^a$. Sensitivity to the $\pm$10 pixels translation shift is discussed in the results.

When multiple misaligned instances are present in an image, the instances are corrected independently. This approach is chosen for two reasons: it allows instances within an image to have varying degrees of translation error and it also enables the network to be robust to incomplete labels with missing instances. Here, a small dataset of images ($x$) and carefully verified ground truth labels ($\upsilon$) are used to train the ACN as shown in Stage 1 of Figure \ref{fig:architecture}.

\subsection{Pointer Segmentation Network}
Assuming $m$ available annotations -- $\upsilon_i^{1}$... $\upsilon_i^m$, where $m < A$ -- common algorithms will struggle to implement Equation \ref{eq:baseline_notation}, as some predicted object instances will not have corresponding true labels for comparison during training. To address this issue, we introduce the PSN, a network that learns to segment an image using only $m$ available annotations. The PSN takes as inputs an image $x_{i}$ and a single channel of points specifying selected instances to be segmented, and it outputs a segmentation mask only for the selected instances. We specify the fraction of instances to be used for training using a parameter $\alpha$, where $\alpha$ is the number of selected instances divided by the number of available instances. Equation \ref{eq:modified_generator_function} shows this formulation, where $p_{i}(\alpha)$ specifies a point within each selected instance, and  $\hat{\upsilon}_i(\alpha)$ denotes the predicted label for instances specified by $p_{i}(\alpha)$:\\
\begin{equation}\label{eq:modified_generator_function} 
\begin{split}
    & f(x_{i}, p_{i}(\alpha))  \rightarrow \hat{\upsilon}_{i}(\alpha) \quad
    \\
    % & \alpha = \frac{num \hfill selected \hfill instances}{num \hfill instances \hfill in \hfill x_{i}}  \\
\end{split}    
\end{equation}

By including a single channel containing points $p_{i}(\alpha)$, our PSN segments only instances that are associated with the points. This offers two benefits: first, we simplify the learning task to specify instances of interest, and second, the network can be trained with common binary cross entropy loss. To handle varying extents of missing annotations, the model is trained by randomly picking $\alpha$ for every image in each epoch; at inference time, all instances of interest are specified using points. 

In the sequential training configuration, the ACN is used to correct a training dataset that is then inputted to the PSN for object segmentation; this process is shown in Stage 2 of Figure \ref{fig:architecture}. Binary cross-entropy loss is used for all networks. Both ACN and PSN use the same baseline architecture (lightUNet) shown in Appendix \ref{lightUnet} and further explained in the results, albeit modified by the number of input channels.

\section{Data}
\label{data}
Three separate datasets are used to train and test the performance of the ACN and the PSN, all described below. During training and testing, we only use images that contain labels.

\subsection{Aerial Imagery for Roof Segmentation}
We use the Aerial Imagery for Roof Segmentation (AIRS) dataset to establish baseline performances for both the ACN and PSN. The AIRS dataset covers most of Christchurch (457$km^2$), New Zealand and consists of orthorectified aerial images (RGB) at a spatial resolution of 7.5 cm with over 220,000 building annotations, split into a training set (T$_{set}$) and a validation set (V$_{set}$). The AIRS dataset provides all building footprints within the dataset coverage area. To mimic more readily-available data, we resample the imagery to 30 cm, an approach which creates imagery more similar to that provided by Google Earth. Next, we slice the resampled images into 128 by 128 pixel patches and discard all patches in which the area occupied by buildings is less than 10 \% of the total area -- this methodology ensures that patches with multiple buildings are selected. Other than this basic filtering, we preserve T$_{set}$ and V$_{set}$. 

\begin{figure}[t]
\begin{center}
\centerline{\includegraphics[width=\columnwidth]{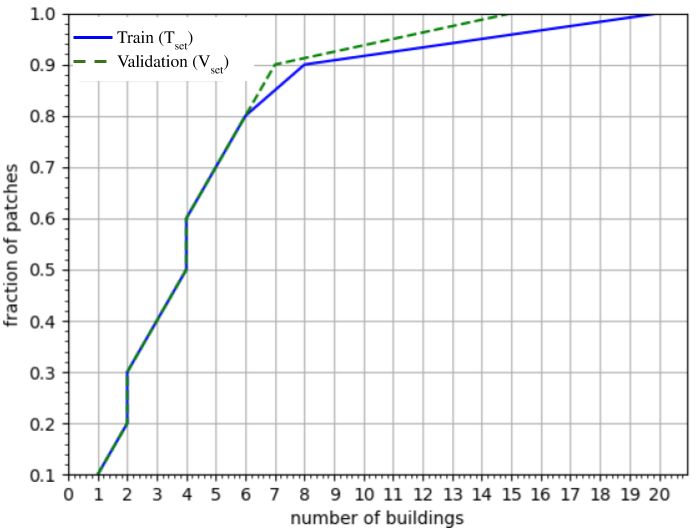}}
\caption{CDF of the number of buildings present in 128x128 patches of the 30cm-resampled AIRS dataset.}
\label{data_fractions}
\end{center}
\vskip -0.2in
\end{figure}

After resampling and filtering, we obtain 99,501 and 10,108 patches from the T$_{set}$ and V$_{set}$, respectively.  We further split T$_{set}$ into 80:20 fractions, where 80\%  is used for training and 20\% for validation. V$_{set}$ is withheld and used as a test set to evaluate performance. Figure \ref{data_fractions} shows the fraction of patches for a given number of buildings in T$_{set}$ and V$_{set}$. Note that some patches contain partial buildings. 

\subsection{OpenStreetMaps}
Humanitarian OpenStreetMaps (OSM), through free, community-driven annotation efforts, provides building footprints by country on their Humanitarian Data Exchange (HDX) platform. While this data provides the best (and only) ground truth for many parts of the world, label quality is highly heterogeneous, both in terms of footprint alignment and coverage. In order to test the performance of the ACN on these incomplete and misaligned building footprints, we pair OSM annotations for Kenya \cite{hdx_kenya} with selected DigitalGlobe tiles from Western Kenya (a box enclosed by 0.176 S, 0.263 S, 34.365 E, and 34.453 E) and closer to Nairobi (a box enclosed by 1.230 S, 1.318 S, 36.738 E, and 36.826 E). The DigitalGlobe tiles have a 50 cm spatial resolution and were collected between 2013 and 2016. Slices measuring 128 by 128 pixels were generated from the DigitalGlobe images, which we then couple with overlapping OSM building labels. We generated human verified ground truth annotations for 500 of the image patches.

\subsection{California Statewide Cropping Map}
We also use crop maps and decameter imagery to demonstrate the flexibility of the PSN.  The California Department of Water Resources provides a Statewide Cropping Map for 2016 \cite{cdwr}; we pair this shapefile with Sentinel-2 satellite imagery to learn to extract crop extents \cite{sentinel}. Red, blue, green, and near-infrared bands -- all at 10m resolution --  are acquired from a satellite pass on August 30, 2016; the bands cover the same spatial extent as Sentinel tile 11SKA (a box enclosed by 37.027 N, 36.011 N, 120.371 W, and 119.112W). Cropped polygons larger than 500m$^2$ are taken from the California cropping map and are eroded by 5m on all sides to ensure that field boundaries are distinct at a 10m spatial resolution. We split the 110km x 110km tile into images patches measuring 128 by 128 pixels and remove all slices that do not cover any cropped areas, leaving a total of 5,681 patches containing an average of 17 fields per patch; these images are split into training, validation, and test sets at a ratio of 60/20/20.

\section{Results}
\label{results}
For all model testing, we report the mean intersection-over-union (mIOU), defined as the intersection of the predicted and true label footprints divided by the union of the same footprints, averaged across the testing dataset.
\subsection{Baseline Model}
We establish the performance of the baseline model (lightUNet) used for both the ACN and PSN by comparing the lightUNet to the UNet architecture proposed by DeepSenseAI \cite{unet_deepsense}. The lightUNet \footnote{See Appendix \ref{lightUnet} for details about the convolutions.} architecture is modified from \cite{unet_deepsense} to perform segmentation with fewer parameters. We refer to the model proposed by \cite{unet_deepsense} as Base-UNet; we train both the Base-UNet and lightUNet models for 30 epochs on the 30 cm resampled AIRS dataset \cite{airs2018}, and we report their mIOU. 
\begin{table}[t]
\centering
\caption{mIOU of Base-UNet\cite{unet_deepsense} and lightUNet for routine segmentation with complete and well-aligned labels. Both models are trained on 30 cm resampled AIRS imagery.}
\begin{tabular}{l r}
{\small\textbf{Models}}
  & {\small \textbf{mIOU}} \\
\midrule
Base-UNet & 0.86 \\
% UNet & 0.853 & 0.924 \\
lightUNet & 0.85 \\
\bottomrule
\end{tabular}
\label{tab:baseline}
\end{table}
Table \ref{tab:baseline} shows that our lightUNet model achieves comparable performance to the Base-UNet when performing routine building segmentation. Our lightUNet model has about half the number of parameters as the Base-UNet and therefore takes less time to train. 

\subsection{Alignment Correction Network}
V$_{set}$ in the AIRS dataset is used to evaluate the performance of the ACN. Random translations were generated between $\pm$ 10 pixels for the xy-axis and applied to ground truth AIRS annotations, resulting in unique translation shifts for each object in an image. The introduction of noise through random translation yields a baseline mIOU of 0.55 for comparison. The shifted annotations together with the images are fed into the ACN, and the corrected annotations are compared to the true annotations to drive the learning process. We report the mIOU on V$_{set}$ when varying amounts of T$_{set}$ data are used for training. Random translations between $\pm$ 10 pixel are applied to all objects in V$_{set}$. When the ACN is trained with 800, 400 and 240 images, the corresponding mIOU on all images in V$_{set}$ are 0.81, 0.77 and 0.67 respectively, compared to the baseline of 0.55. This suggests that the ACN performs better when more images are used but can learn with only a couple hundred training images.
% Random translations were generated in 5 pixel intervals between 0 and 20 pixels in both the x- and y-axes, and then applied to ground truth annotations; this process results in unique shifts for each object in an image. The images and the shifted annotations are inputted to ACN, which learns to correct the shifted annotations using the true labels. In all tests, the ACN is trained with 400 images. In Table \ref{tab:random_translated}, we present mIOU for both shifted labels before and after ACN correction, demonstrating that the ACN improves label accuracy for all translations shifts. 
\begin{table}[h]
\centering
\caption{mIOU before and after ACN correction.}
\begin{tabular}{l | c c }
 & \multicolumn{2}{c}{mIOU} \\
 \hline
{\small\textbf{Translation Shift ($\pm$ pixels)}}  & {\small \textbf{Before ACN}} & {\small \textbf{After ACN}}  \\
\midrule
0 to 5 & 0.63 & 0.81 \\
5 to 10 & 0.40 & 0.73 \\
10 to 15 & 0.26 & 0.46 \\
15 to 20 & 0.18 & 0.28 \\
\bottomrule
\end{tabular}
\label{tab:random_translated}
\end{table}

Using the ACN model trained with 400 images and random translation shifts between $\pm$ 10 pixels, we evaluate the robustness of the ACN to varying levels of translation shifts.  Table \ref{tab:random_translated} shows mIOU before  and after ACN correct, when different ranges of translations shifts are applied to  V$_{set}$. Across all translation shifts the ACN is able to perform some realignment of annotations, even for translations (>10 pixels) which the model was never trained on.
% To better inform the human-verified annotation collection process, we next investigate the question of how many annotations per image are required for training. If a single annotation per image is sufficient, the annotation acquisition task is greatly simplified and the segmentation cost can be further reduced. To explore this question, we train the ACN with one repeated building annotation per image, referred to as the `repeated instance' training configuration. In this configuration, we pre-select a single object instance per image and its corresponding annotation, and we use this same annotation in all training epochs. In the `random instances' training configuration, the ACN uses a randomly-selected building instance for each image during every training epoch; accordingly, this configuration requires multiple instance annotations in a given image. Here, there is no guarantee that the same instance is selected during the next epoch. Table \ref{tab:misaligned} shows that multiple instances per image improves the annotation correction process, although the gains are minimal. 

\begin{figure}[h]
\begin{center}
\centerline{\includegraphics[width=\columnwidth]{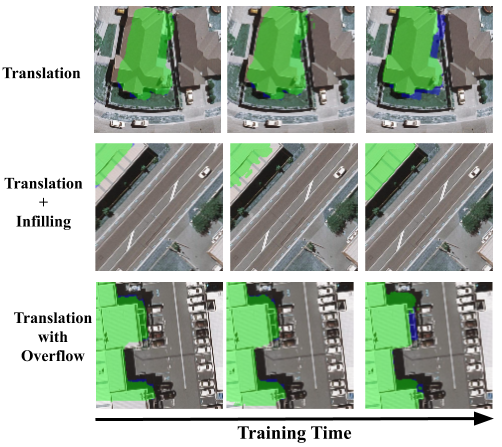}}
\caption{Types of annotation corrections performed by the ACN when trained with 800 images. \textcolor{green}{Green shows corrected annotations}. \textcolor{blue}{Blue shows misaligned annotations.}}
\label{realignment}
\end{center}
% \vskip -0.2in
\end{figure}

We observe two types of alignment correction as outputs from the ACN: translations and translations plus infilling. Infilling occurs when the misaligned annotation area is less than the building area. In the translation plus infilling case, the model both shifts the annotation and fills the missing portion of the annotation. Overflow is sometimes observed upon correcting the label, resulting in the corrected annotation exceeding the building outline. Figure \ref{realignment} shows examples of both types of corrections when training on 800 images. This figure demonstrates how the ACN learns over time: green outlines show predictions from the ACN and blue outlines show misaligned annotations which the ACN takes as input.

\subsection{Pointer Segmentation Network}
As an alternative to traditional segmentation models, we propose the Pointer Segmentation Network (PSN), a network that takes in an additional channel with points of interest and returns a single channel output with annotations. The PSN was evaluated separately from the Alignment Correction Network (ACN); this section focuses on reporting segmentation performance on the AIRS dataset when partial -- but well-aligned -- labels are used.  To appropriately compare the PSN with the lightUNet, we evaluate model performance using all annotations in every image of V$_{set}$. Here, we compare the ability of both networks to segment every building instance in the image, having learned with missing annotations. Table \ref{tab:table2} reports the performance of the lightUNet and the PSN with varying fractions of selected annotations ($\alpha$): As $\alpha$ decreases, performance of the PSN remains robust, indicating that the network still learns the segmentation task despite missing annotations. By specifying the points of interest, the PSN outperforms the lightUNet model.

Table \ref{tab:table2} also presents results for two different methods of acquiring the required building points: using building centroids versus using a randomly generated point from within the corresponding annotation. By comparing the performance of the PSN using centroids with that of randomly generated points, the best annotation strategy to be used at inference can be determined. We find that the PSN performs better when centroids are used to train the model: This suggests that annotators should strive to extract points near the center of buildings to ensure better segmentation outcomes during inference. Additionally, because the extent of missing annotations may not be known \textit{a priori} for datasets, we evaluate how the network handles heterogeneous (Het.) amounts of label completeness by sampling $\alpha$ from a random uniform distribution between 0 and 1. The uniform distribution ensures an equal chance for alpha to take on any value between 0 and 1. $\alpha$ is resampled for each image during every training epoch. Table \ref{tab:table2} shows that the PSN remains robust at performing segmentation and works for a heterogeneous $\alpha$ that varies across images. Although $\alpha$ will likely differ across images but remain constant for a given image at a particular time, during training we allow $\alpha$ to change over every training epoch for a given image, enabling our approach to be robust against images taken at different times where new construction may have occurred.
\begin{table}[t]
\centering
\caption{mIOU of PSN and lightUNet for all buildings in V$_{set}$ images, when trained with varying $\alpha$.}
\begin{tabular}{l l r}
{\small\textit{}}
 & & {\small \textbf{mIOU}} \\
\midrule
$\alpha$ = 1 & PSN (centroid) & 0.90 \\
 & lightUNet (centroid) & 0.85  \\
\midrule
$\alpha$ = 0.7 & PSN (centroid) & 0.89  \\
     & PSN (non-centroid) & 0.83 \\
 & lightUNet (centroid) & 0.53 \\
\midrule
$\alpha$ = 0.5 & PSN (centroid) & 0.87 \\
 & lightUNet (centroid) & 0.18  \\
 \midrule
$\alpha$= Het. & PSN (centroid) & 0.87 \\
 & lightUNet (centroid) & 0.71 \\
 \bottomrule
\end{tabular}
\label{tab:table2}
\end{table}

\begin{figure}[h]
\begin{center}
\centerline{\includegraphics[width=\columnwidth]{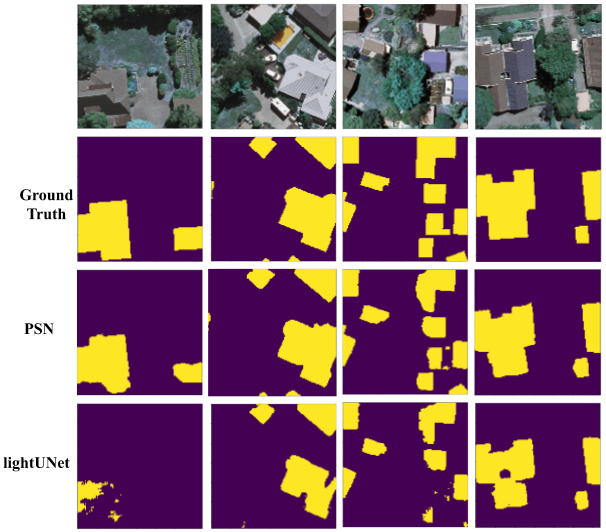}}
\caption{Annotations from PSN and lightUNet models when trained with ${\alpha=0.7}$. Predictions are made for all building instances in the image and are compared to the ground truth.}
\label{psn_performance}
\end{center}
% \vskip -0.2in
\end{figure}

Figure \ref{psn_performance} shows how the PSN learns -- and where non-PSN type networks fail -- when learning with missing annotations. The figure shows some outputs of the PSN and the lightUNet model when both are trained with ${\alpha=0.7}$ and used to predict all building instances present within the image. Although both networks are trained with missing annotations, generated annotations from the PSN are more visually accurate.

\subsection{Sequential Testing}
The AIRS dataset is used to evaluate the sequential performance of our two-stage methodology shown in Stage 2 of Figure \ref{fig:architecture}, whereby the ACN and PSN are trained and tested sequentially. Using T$_{set}$, we establish two training datasets for the sequential process: T1, containing misaligned labels generated from the true T$_{set}$; and T2, containing ACN-corrected T1 labels. The ACN model trained with 400 training images is used to generate T2. The noise present in both training datasets is captured by the mIOU listed in Table \ref{tab:sequential}. The PSN and lightUNet models are trained on T1 and T2 using ${\alpha=Het}$ with an identical implementation of label withholding to that described in the previous section. The trained models are used to segment V$_{set}$ images; we compare predicted annotations to the true annotations to attain the performance metrics reported in Table \ref{tab:sequential}. 

Table \ref{tab:sequential} shows that, with $\alpha=Het$, the PSN performs significantly better than the lightUNet when trained on either misaligned labels (T1) or ACN-corrected labels (T2). Again, we find that with incomplete labels, regardless of alignment quality, the PSN outperforms the lightUNet. Moreover, in both training configurations, PSN mIOU performance nears that of the training dataset. As a result, we conclude that the PSN is able to predict object extents at a similar accuracy to that of the training dataset.
\begin{table}[h]
\begin{center}
\caption{Performance of the segmentation architectures. The ACN is trained with 400 images; both segmentation networks are trained with ${\alpha=Het.}$ available annotations.}~
\label{tab:sequential}
\begin{tabular}{l r r }
{\small\textit{}}
  & {\small \textbf{mIOU}} \\
\midrule
T1: Misaligned train dataset  & 0.57  \\
% V1: Misaligned training dataset  & 0.5  & 0.74  \\
PSN (trained on T1)             & 0.54      \\
lightUNet (trained on T1) & 0.17     \\
\midrule
T2: ACN-corrected train dataset   & 0.81   \\
% V2: ACN-Corrected training dataset   & 0.77  & 0.87  \\
PSN (trained on T2)       & 0.79    \\
lightUNet (trained on T2)  & 0.74    \\
\bottomrule
\end{tabular}
\vskip -0.2in
\end{center}
\end{table}

\begin{figure}[t]
\begin{center}
\centerline{\includegraphics[width=\columnwidth]{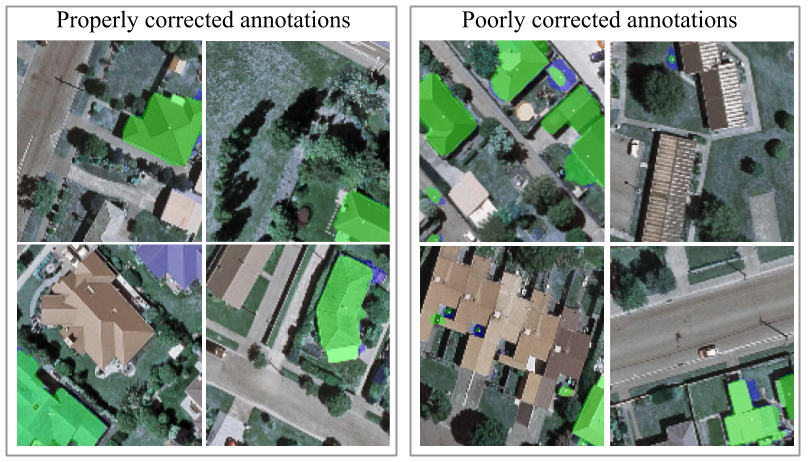}}
\caption{Sample images showing PSN performance when trained with corrected annotations. \textcolor{blue}{Blue footprints} show ACN-corrected annotations. \textcolor{green}{Green footprints} show PSN-predicted annotations trained with ${\alpha=Het.}$ and 400 ACN-corrected labels. PSN performance is dependent on the quality of corrected annotations.}
\label{end_to_end_performance}
\end{center}
% \vskip -0.2in
\end{figure}

Figure \ref{end_to_end_performance} presents outputs from the PSN when trained with ACN-corrected annotations: corrected annotations from the ACN are shown in blue and predicted outputs from the PSN are shown in green. In the left half of Figure \ref{end_to_end_performance}, we present properly corrected ACN-labels and demonstrate that the PSN is able to predict building footprints accurately when corrected annotations are accurate. The right half of the figure shows poorly corrected annotations: These corrected annotations fall on roads, grass, or across the actual building extent. In these cases, the PSN tries to predict a building footprint where there is no building. Accordingly, we conclude that improvements to the ACN can further improve PSN performance, as more accurate training labels will allow for better label prediction. Nonetheless, in the presence of misaligned annotations and partial labels, we are able to achieve better performance with our sequential architecture than with traditional segmentation approaches.

\subsection{ACN Application: Realignment of OSM Annotations}
In many parts of the world, ground truth is rare or nonexistent; moreover, what resources do exist often have significant accuracy issues. Despite potential shortcomings, these datasets can provide unique insight into conditions on the ground, and if their quality can be improved, they offer immense value to researchers. To confirm the performance of our realignment method on noisier images and labels, we tested the ACN on OSM building polygons in Kenya, a dataset containing considerable amounts of label misalignment. Of the 500 human-verified ground truth image labels generated for Kenya, 400 are used to train the ACN and 100 to validate. The extent of noise in OSM labels is measured by comparing the labels to the human-verified ground truth labels. mIOUs of 0.30 and 0.31 for the train and validation data respectively were recorded, when comparing OSM labels to their ground truth counterparts. OSM training labels are used to train the ACN and the trained model is ran on the 100 validation labels. A 50 \% improvement in mIOU from 0.31 to 0.47 is observed on the 100 validation images. This suggests that our approach is transferable to open source labels and offers gains even with noiser images and labels, using a small dataset.
% By comparing these values to those in Table \ref{tab:random_translated}, we can see that the average HotOSM label has a misalignment of approximately $\pm$10 pixels in both x- and y- directions before correction.
% \begin{table}[h]
% \centering
% \caption{mIOU for HotOSM labels in training and validation dataset, as compared to hand-labelled annotations.}
% \begin{tabular}{l | c c }
%  & \multicolumn{2}{c}{mIOU} \\
%  \hline
%   {\small \textbf{Dataset}} & {\small \textbf{Before ACN}} & {\small \textbf{After ACN}}  \\
% \midrule
% Training (400 images)   & 0.30 & 0.47 \\
% Validation (100 images)  & 0.31 & 0.47 \\
% \end{tabular}
% \label{tab:hotosm}
% \end{table}
\begin{figure}[h]
\begin{center}
\centerline{\includegraphics[width=\columnwidth]{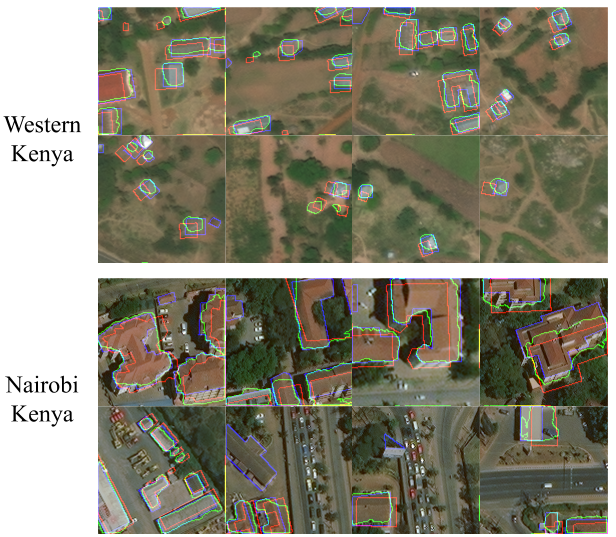}}
\caption{\textcolor{blue}{Hand-labelled annotations}, \textcolor{red}{OSM annotations} and \textcolor{green}{ACN-corrected annotations}. The ACN is trained on 400 images from Western Kenya and Nairobi, and improves label quality despite the noisier training data.}
\label{osm_performance}
\end{center}
% \vskip -0.2in
\end{figure}

Figure \ref{osm_performance} shows a sampling of ACN-corrected OSM annotations for images in the validation dataset: Hand-labelled annotation are shown in blue, OSM annotations are shown in red and corrected annotations are shown in green. Overall, we find that the ACN is able to correct misaligned OSM annotations both in rural and urban regions. In rural Western Kenya, where buildings tend to be smaller, the ACN shifts OSM footprints to better align with the buildings. We observe that the noisier image quality makes it more difficult for the ACN to identify extremely small buildings. In more urbanized Nairobi, the ACN also improves the alignment of OSM annotations, albeit with some failure cases. 

%Occasionally, OSM footprints cover both the rooftop and the side of the building: In these cases where the quality of the annotation is more ambiguous, the corrected annotations looks very similar to their OSM counterparts.

\subsection{PSN Application: Cropland Segmentation}
Next, we apply the PSN to the task of cropland segmentation using Sentinel-2 imagery and a 2016 California cropping map. Knowing exact field outlines provides valuable information to farmers, planners, and governments; however, a lack of reliable, location-specific ground truth often hampers these efforts. We demonstrate the ability to accurately learn cropland extents using only a subset of fields, instead of requiring the comprehensive set of training polygons that would be necessary for traditional segmentation networks. Similar to previously described tests, we quantify the performance of the PSN in recreating these field boundaries as we select a certain fraction of the annotations, comparing results to those of the lightUNet. Table \ref{tab:ag} presents these results.
\begin{table}[b]
\centering
\caption{mIOU for all field boundaries in test set, for varying $\alpha$ values.}
\begin{tabular}{l l r r }
{\small\textit{}}
 & & {\small \textbf{mIOU}} \\
\midrule
$\alpha$ = 1 & PSN & 0.92  \\
 & lightUNet & 0.75  \\
\midrule
$\alpha$ = 0.75 & PSN & 0.91 \\
 & lightUNet & 0.69 \\
% \midrule
% $\alpha$ = 0.5 & PSN & 0.90  \\
%  & lightUNet & 0.00 \\
 \bottomrule
\end{tabular}
\label{tab:ag}
\end{table}

\begin{figure}[t]
\begin{center}
\centerline{\includegraphics[width=\columnwidth]{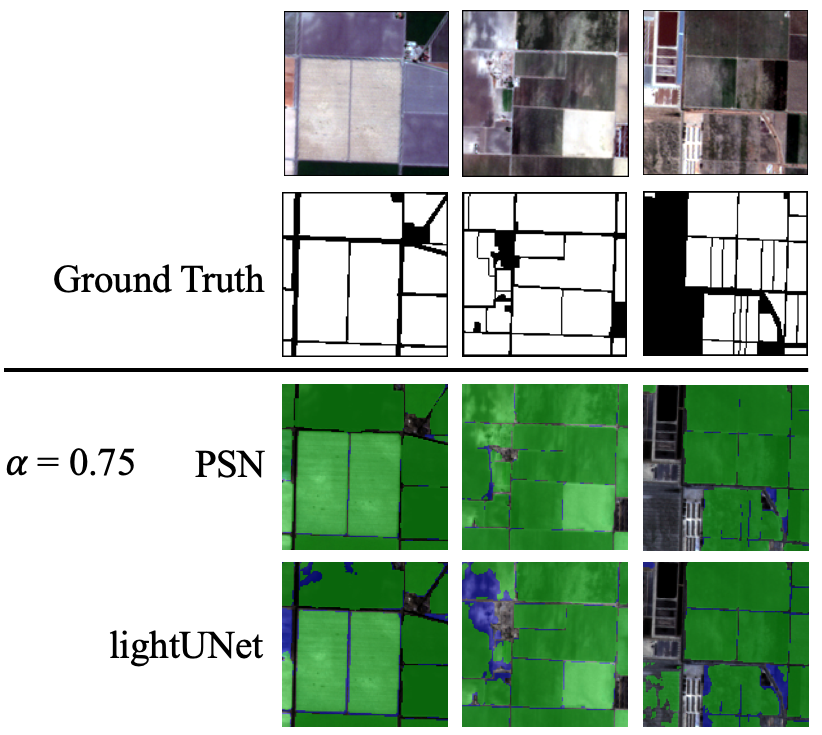}}
\caption{Sample images and ground truth labels showing cropland extent in California; also shown in green are \textcolor{green}{PSN and lightUNet predicted footprints} $\alpha = 0.75$, overlaid on \textcolor{blue}{true cropland polygons}, shown in blue. PSN predictions remain highly accurate. Comparatively, the lightUNet predicts only a portion of the crop extents correctly}
\label{ag_img}
\end{center}
\vskip -0.2in
\end{figure}

At all fractions of available training data shown in the table, the PSN outperforms the lightUNet in segmenting croplands. After 40 training epochs, the PSN is able to predict all field boundaries for the test set across both values of $\alpha$. When trained with all annotations ($\alpha$ = 1), the PSN achieves a mIOU of 0.92. In contrast, the lightUNet only reaches a mIOU of 0.75 when $\alpha$ = 1, and sees its performance significantly diminish as field boundaries are withheld.
% with $\alpha$ = 0.5, this metric drops only to 0.90. : With only 50\% of available annotations, the lightUNet mIOU performance degrades to 0. 
Figure \ref{ag_img} shows the PSN- and lightUNet - recreated field polygons when the models are trained with $\alpha$ = 0.75 and are asked to predict all polygons within an image. The true cropland polygons are shown in blue while the predicted polygons are shown in green; all examples shown come from the test set.

% Figure \ref{ag_img} shows the PSN- and lightUNet - recreated field polygons when the models are trained with $\alpha$ = 0.75 and $\alpha$ = 0.5 and are asked to predict all polygons within an image. The true cropland polygons are shown in blue while the predicted polygons are shown in green; all examples shown come from the test set.
% \begin{figure}[t]
% \begin{center}
% \centerline{\includegraphics[width=\columnwidth]{figures/preds_compiled.png}}
% \caption{Sample images and ground truth labels showing cropland extent in California; also shown in green are \textcolor{green}{PSN and lightUNet predicted footprints} for two levels of available annotations ($\alpha = 0.75$, $0.5$) overlaid on \textcolor{blue}{true cropland polygons}, shown in blue. At both $\alpha$ values, PSN predictions remain highly accurate. Comparatively, the lightUNet predicts a portion of the crop extents correctly at $\alpha=0.75$ and cannot predict any of the labels when trained with $\alpha=0.5$.}
% \label{ag_img}
% \end{center}
% \vskip -0.2in
% \end{figure}
These results demonstrate the viability of the PSN in delineating field boundaries and the preferability of our method over a baseline alternative, when  the acquisition of field boundaries is expensive. In locations with low data availability and smaller, non-uniform field boundaries, the PSN provides a reliable method for determining cropped area polygons.

\section{Conclusion}
\label{conclusion}
As the demand for extracting information from satellite imagery increases, the value of reliable, transferable object segmentation methodologies -- especially ones that compensate for noise and inaccuracies in training data -- increases in parallel. In this paper, we present a novel and generalizable two-stage segmentation approach that address common issues in applying deep learning approaches to remotely-sensed imagery. First, we present the Alignment Correction Network (ACN), a model which learns to correct misaligned object annotations. We test the ACN on a set of alignment errors, including i) misalignment of the AIRS dataset, ii) existing and substantial misalignment errors within the OSM Kenyan building footprint dataset. Overall, we find that the ACN significantly improves annotation alignment accuracy.

We also introduce the Pointer Segmentation Network (PSN), a model which reliably predicts an object's extent using only a point from the object's interior. The value of the PSN lies in learning to segment objects within an image despite incomplete or missing annotations, an issue which both hinders traditional segmentation efforts and is common in many ground-truth datasets. We train and test the PSN on the AIRS dataset and find that the model can accurately predict building extent regardless of the fraction of available annotations present or where the training point resides within the object. We also evaluate the performance of the PSN for cropland segmentation using Sentinel imagery and a 2016 California cropland map as inputs, demonstrating that the model can reliably learn cropland polygons regardless of the fraction of available annotations.  Overall, for all testing configurations -- those which vary the fraction of available training annotations and those which change the location of where the training point lies-- and for both object segmentation applications presented -- building footprint and cropland extent predictions -- the PSN outperforms a baseline segmentation model.

Lastly, we sequentially link the ACN and PSN to demonstrate the ability of the combined networks to accurately segment objects having learnt from misaligned and incomplete training data. Taken together, we envision our proposed networks providing value to the community of researchers and scientists looking to extract information from widely-available satellite imagery and unreliable ground-truth datasets.

% \begin{acks}
% \end{acks}

\bibliographystyle{ACM-Reference-Format}
%\bibliography{sample-base}

\newpage
\appendix
\section{Architecture}
\label{lightUnet}
\begin{figure*}[b]
  \centering
  \includegraphics[width=2\columnwidth]{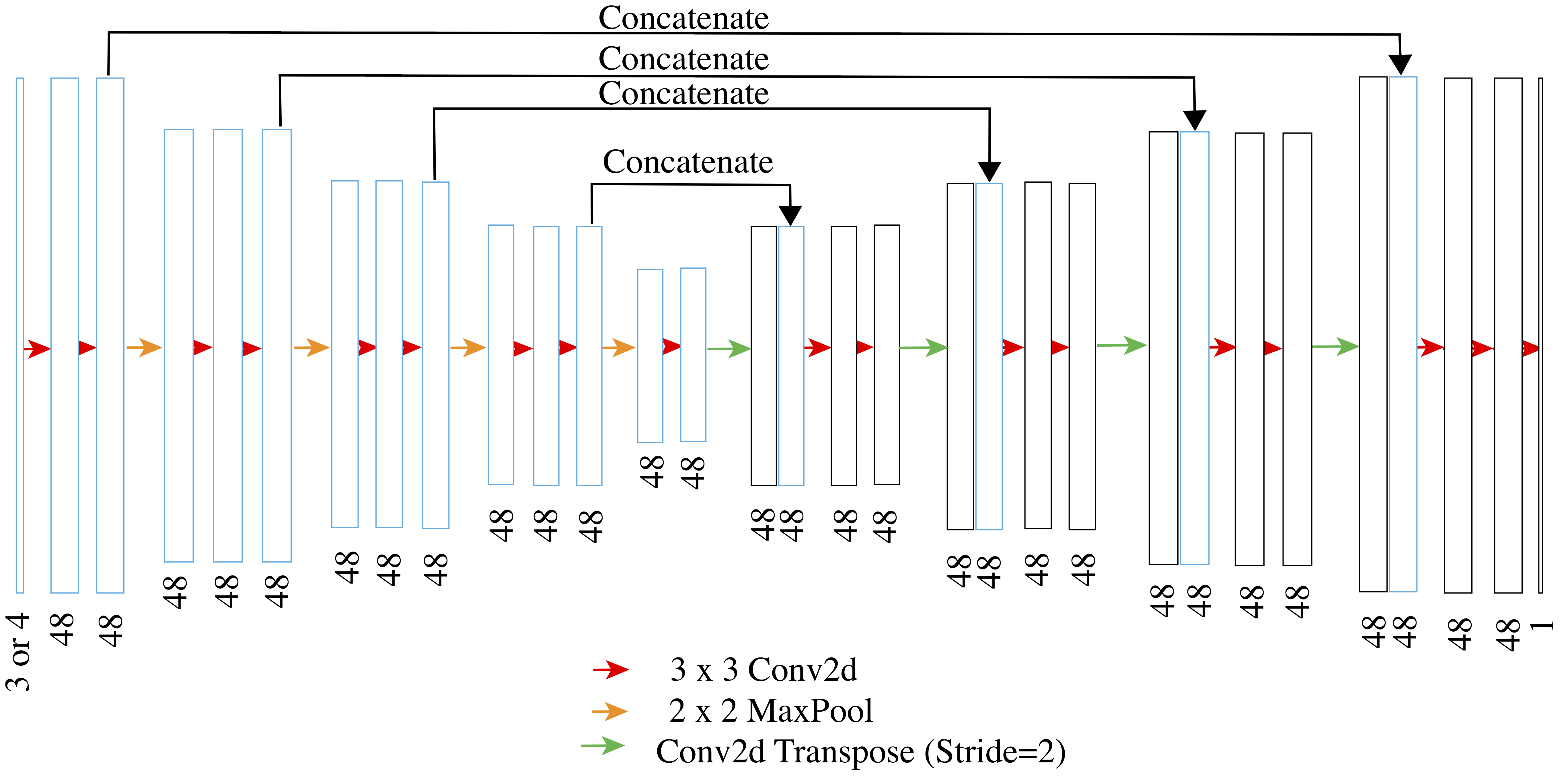}
  \caption{Architecture used for both the Alignment Correction Network (ACN) and the Pointer Segmentation Network (PSN). Four input channels are used for both ACN and PSN, while three are used for the lightUNet. This network is modified from \cite{unet_deepsense} by reducing the number of filters to 48 and maintaining the same filter size through out the network. In addition, the network uses dropout in addition to batch normalization after every epoch.}~\label{fig:lightunet}
\end{figure*}
\end{document}